\documentclass{article}
\usepackage{acra}
\usepackage{amsmath}
\usepackage[capitalise,noabbrev]{cleveref}
\usepackage[free-standing-units=true]{siunitx}
\usepackage{graphicx}
\usepackage{balance}
\usepackage{fancyhdr}
\usepackage{ragged2e}

\fancypagestyle{firstpage}
{
    \fancyhf{}

    \fancyhead[L]{\small \centering \noindent Preprint accepted at the Australian Conference on Robotics and Automation (ACRA 2022), Brisbane, Australia, 2022. 
    }

}

\title{
Introspection-based Explainable Reinforcement Learning\\in Episodic and Non-episodic Scenarios
}
\author{
Niclas Schroeter  \\ 
Universität Hamburg \\
Germany \And 
Francisco Cruz \\ 
UNSW Sydney \\
Australia \And
Stefan Wermter \\
Universität Hamburg \\
Germany}

\begin{document}

\maketitle

\begin{abstract}
With the increasing presence of robotic systems and human-robot environments in today's society,
understanding the reasoning behind actions taken by a robot is becoming more important. To increase this
understanding, users are provided with explanations as to why a specific action was taken. Among other
effects, these explanations improve the trust of users in their robotic partners. One option for creating
these explanations is an introspection-based approach which can be used in conjunction with reinforcement
learning agents to provide probabilities of success. These can in turn be used to reason about the actions
taken by the agent in a human-understandable fashion. In this work, this introspection-based approach is
developed and evaluated further on the basis of an episodic and a non-episodic robotics simulation task.
Furthermore, an additional normalization step to the Q-values is proposed, which enables the usage of the
introspection-based approach on negative and comparatively small Q-values. Results obtained show the
viability of introspection for episodic robotics tasks and, additionally, that the introspection-based
approach can be used to generate explanations for the actions taken in a non-episodic robotics environment as
well.
\end{abstract}

\section{Introduction}
\label{Introduction}

Over recent years, explainability in robotic systems has gained traction as a new field of
research~\cite{KABLSSGWW22}~\cite{sado2022explainable}.
Explainable AI (XAI) enables the robotic system to reason about its actions or justify them in some manner~\cite{anjomshoae2019explainable}, giving humans the ability to understand why the AI behaves in the observed fashion. 
This understanding can, in turn, increase the trust of humans in robotic systems,
which is especially important in high-risk environments, as better understanding and trust in the system can increase the overall performance of these human-robot systems~\cite{wang2016trust}. 
It is important to distinguish between explainablity and interpretability, as defined in \cite{rosenfeld2019explainability}. 
While interpretability provides a representation of the underlying logic of an agent, explainability aims to provide reasoning in a human-understandable fashion.

Since users in these human-robot systems are not guaranteed to be experts on the topic of AI, it is important to generate explanations that can be understood by both expert and non-expert users. 
In order to realize this, the explanations need to forgo technical terms wherever possible. 
As an example, providing the different Q-values would not help a non-expert user in understanding the reasoning behind an AI's decision. Instead of using technical terms that non-experts are not familiar
with, a concept that can be understood by most people should be used instead, for example the
concept of success probabilities. Recent work has shown that stand-alone or counterfactual
explanations based on probabilities for actions in different robotic scenarios are perceived as more
useful by non-experts than similar explanations that rely on technical terms like
Q-values~\cite{cruz2022evaluating}. Alongside other approaches, goal-driven
XAI~\cite{sado2022explainable} focuses on generating explanations that are understandable by
non-practitioners.

There are many different approaches and algorithms for generating these explanations, such as the
memory-based and the learning-based approaches, highlighted among others in \cref{sec:rel}. 
Our approach focuses on generating goal-driven explanations for reinforcement learning (RL) agents using the introspection approach suggested in \cite{cruz2021explainable}.
When applying this algorithm, the underlying Q-values that govern a decision are transformed into probabilities of success for a certain action, which can then be used to reason about the action taken by the agent. 
In order to evaluate this approach, a robot simulation is built which involves a flying agent that looks for an object with random spawn positions, upon which an episodic and a non-episodic task are based. 
The introspection-based approach is then applied to these tasks to generate explanations in the form of success probabilities for the actions the flying agent takes.

\section{Related Work}
\label{sec:rel}
Many different approaches to explainability are currently being investigated. 
As networks grew deeper over the recent years, efforts to understand the decision they make have resulted in different visualization techniques, for example in different saliency methods that visualize decision-relevant regions of the input, such as Grad-CAM++~\cite{chattopadhay2018grad}. 
Rupprecht~et~al. apply novel visualization techniques to explain decisions made by an RL agent and to
subsequently find flaws in their behaviour~\shortcite{rupprecht2019finding}. 
In~\cite{dazeley2021explainable}, the authors reason that explainable reinforcement learning (XRL)
should be considered separate from interpretable machine learning, as the field of XRL allows for much
more than just interpreting decisions. Alongside the introduction of the conceptual Causal XRL Framework
that aims to unify XRL research as well as the usage of RL in general to develop Broad-XAI, they also
surveyed current literature, identifying different branches regarding XRL, for example the usage of
introspection in value-based RL methods, which this work resides in, as it demonstrates the
transformation of Q-values into probabilities that can be embedded into standalone or counterfactual
explanations.

Looking at XRL in specific, many approaches utilize technical
explanations to achieve explainability for the corresponding agent~\cite{puiutta2020explainable}.
Some approaches integrate the generation of human-readable explanations into different RL frameworks.
Shu~et~al. have integrated the generation of human-readable instructions into their
multi-task RL framework by tying them to skill acquisition, which can also be used to understand the
agent's behaviour~\shortcite{shu2017hierarchical}, whereas Verma et al. represent the learned policies in a high-level domain language~\shortcite{verma2018programmatically}. In \cite{hein2018interpretable}, the explanations are generated using a
hybrid approach that involves genetic programming alongside RL. The resulting explanations consist of
algebraic equations though, which leads to limited readability for humans. The approach shown in
\cite{rietz2022hierarchical} utilizes hierarchical goals as context for the explanations of single
actions to increase interpretability in scenarios with multiple optimal policies, relying on
hierarchical reinforcement learning to create these explanations.

Explanations can also be expressed in terms of probabilities of success. The memory-based approach first
proposed in \cite{cruz2019memory} utilizes an episodic memory to generate these probabilities, however
managing this memory increases the space complexity considerably, which hinders this approach from being
applied to scenarios with continuous states. Alongside the introspection-based approach, a
learning-based approach was also suggested in \cite{cruz2021explainable}, in which the success
probabilities are learned in parallel to the actual training. Both the learning-based and the
intro-spection-based approach appear to be on equal footing in the evaluation of the results, although
the introspection requires fewer resources. In addition to this, further evaluations of the
introspection approach show that it is also applicable to simulated non-episodic, continuous
tasks~\cite{ayala2021explainable}\cite{portugal2022analysis}.

The goal of this work is to provide further evaluations of the introspection-based approach to cement
its viability as an approach to generate explanations of the RL agent's behaviour. In particular, the
approach has not been tested on robotic non-episodic, continuous tasks yet.

\section{Methods}
\label{sec:methods}
\subsection{Reinforcement Learning (RL)}
\label{sub:rl}
In reinforcement learning, an agent discovers which actions yield the highest reward in an environment
by trying them and observing the effect on the environment~\cite{sutton2018reinforcement}. Formally, RL
problems are represented by Markov decision processes (MDP). MDPs consist of sets of states \textit{S},
actions \textit{A} and rewards \textit{R}. Depending on the current state, an action can be taken, which,
in return, yields a certain reward. The goal of the agent is to learn an optimal policy
$\pi^* : S \rightarrow A$, in which the highest possible reward is produced from a given state,
alongside an optimal action-value function $q^*(s,a) = \max\limits_{\pi} q^\pi (s,a)$. In order to find
$q^*$, the Bellman optimality equation shown in \cref{eq1} is solved. In this equation, $s$ represents
the current state of the agent, $a$ is the action performed, $r$ is the reward that is received after
performing $a$ to reach state $s'$, $\gamma$ is the discount rate and $p$ is the probability of reaching
state $s'$ from $s$ when the action taken is $a$.

 \begin{equation}\label{eq1}
q^*(s,a) = \sum_{s',r}{p(s', r | s, a)[r + \gamma \max\limits_{a'} q^*(s', a')]}
\end{equation}

\subsection{Generating explanations}
\label{sub:expl}
The probabilities that are used in the explanations of the agent's behaviour can be generated in different manners, depending on the utilized approach. 
For the introspection-based approach, the success probability $\hat{P}_s$ is estimated directly from the Q-values. 
Due to this, no additional memory is needed to calculate the probabilities, as it would be the case with both the memory-based and the learning-based approaches. 
The final transformation is shown in \cref{eq2} and is explained in more depth in
\cite{cruz2021explainable}. To derive the equation for the transformation, the general approach of
temporal difference learning needs to be examined. Focusing on only episodic tasks for now, any
Q-value $Q(s,a)$ represents the potential future reward, which, in a simplified manner, can be
considered to be the repeated multiplication of the discount factor $\gamma$ and the terminal reward
$R^T$. This can be expressed as $Q(s,a)\approx R^T \cdot \gamma^n$. Once the Q-values converge to their true values, they can be used to estimate the distance $n$ to
the terminal reward, which was introduced in the previous term:

\begin{equation}\label{eq1.5}
n \approx log_{\gamma}\frac{Q(s,a)}{R^T}
\end{equation}

After obtaining the estimated distance $n$, a logarithmic base transformation is applied to estimate
a success probability, alongside a multiplication with $\frac{1}{2}$ and the addition of $1$ to shift
the curve to an area where the probabilities are more plausible. Finally, the values are rectified to
the interval $[0,1]$ to accommodate for any values that would lie outside of the plausible range for
probabilities, leading to the final form of the transformation equation, as shown in \cref{eq2}.

In this equation, $Q^{*}(s,a)$ refers to the Q-values and $\sigma$ is used to account for stochastic
transitions.
$R^M$ refers to the maximum reward, which differs between episodic and non-episodic tasks. 
For episodic tasks, the maximum reward is the terminal reward $R^T$, whereas the maximum reward in a non-episodic task is the highest reward that can be obtained in any single time step~\cite{ayala2021explainable}, denoted as $R^S$. 
Since this transformation only relies on the Q-values and a maximum reward $R^M$, any method that is built on the calculation of Q-values can utilize this approach for the generation of explanations in the form of success probabilities, such as SARSA~\cite{rummery1994line} or DQN~\cite{dqn}.

Furthermore, using only the Q-values alongside the transformations enables the user to query the success probabilities of certain actions at any point in time, beginning at the start of training. 
However, it should be noted that the success probabilities will not converge until the Q-values have as well.

\begin{equation}\label{eq2}
\hat{P}_s \approx \Big[(1 - \sigma) \cdot \Big(\frac{1}{2} \cdot \log_{10}\frac{Q^*(s,a)}{R^M} + 1\Big)\Big]^{\hat{P}_s \leq 1}_{\hat{P}_s \geq 0}
\end{equation}

\section{Experiments}
\label{sec:exp}

\begin{figure}[h]
	\centering
	\includegraphics[scale=0.2]{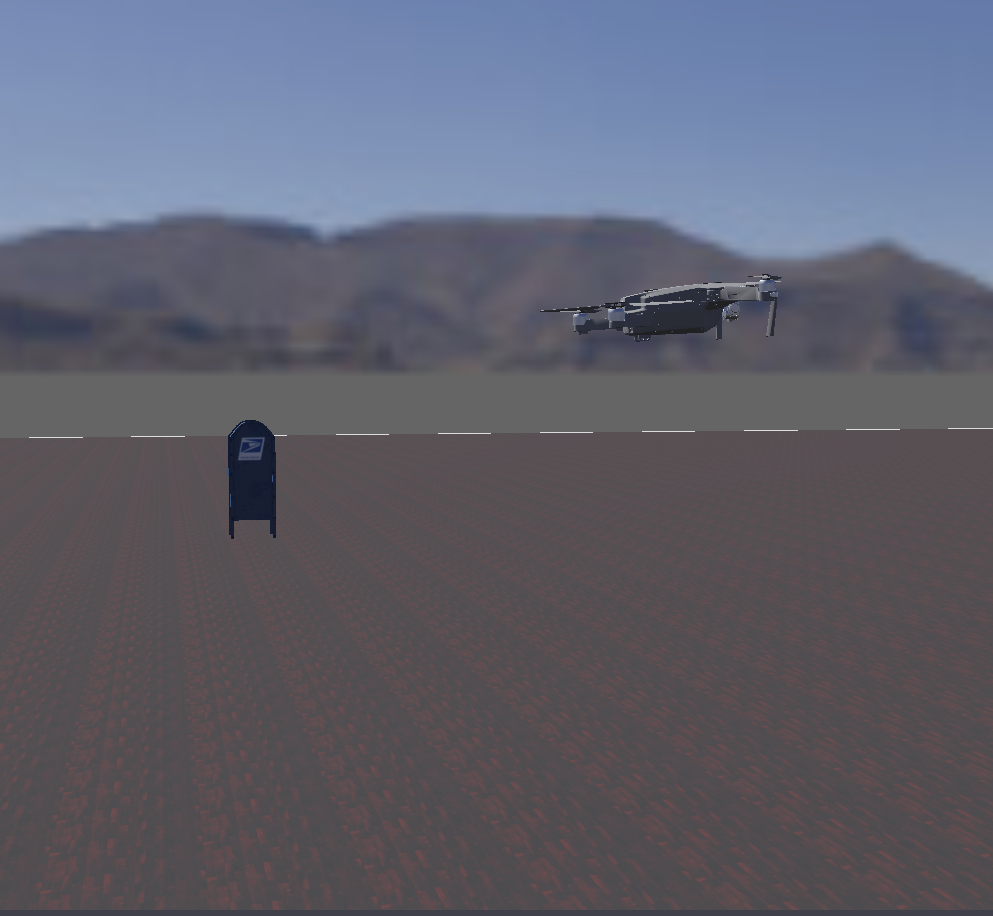}
	\caption[Robot Simulation]{Screenshot of the running simulation. The drone in front is looking for the mailbox in the background.}
	\label{fig:sim}
\end{figure}
\begin{figure}[h!]
	\centering
	\includegraphics[scale=0.3]{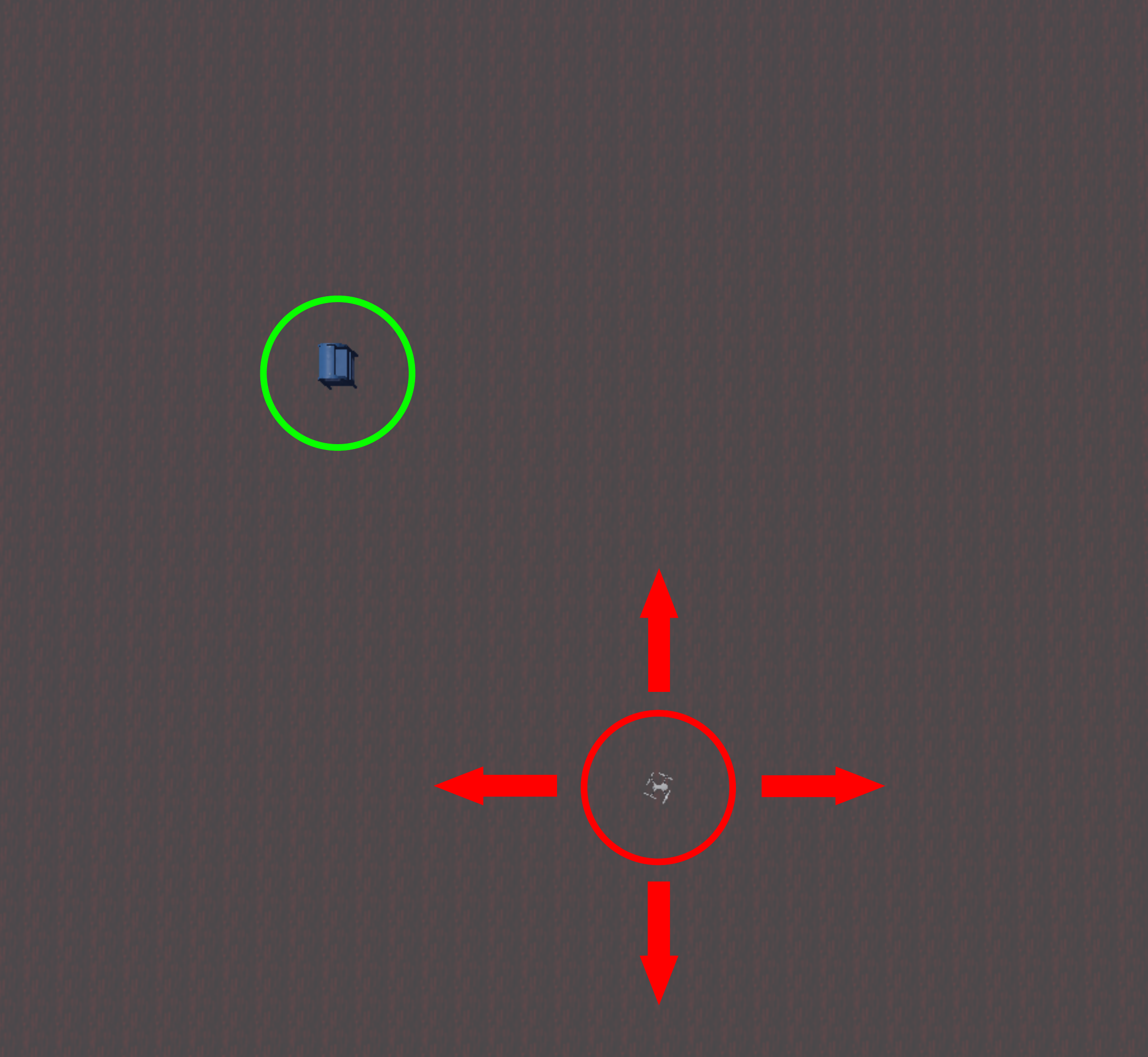}
	\caption[Movement Options]{Top-down view of the simulation. The drone (in the red circle) can move
	in any of the four cardinal directions, as visualized by the arrows, to close the distance to the
	mailbox (in the green circle).}
	\label{fig:movement}
\end{figure}

In order to evaluate the generation of explanations with the help of introspection, a simulation
environment for robotics is used. There are many options to choose from, and many of them have been
compared in different studies~\cite{pitonakova2018feature}\cite{ayala2020comparison}. For this work,
Webots~\cite{webots} will be used, as it has moved to an open source model recently and is also
comparatively efficient regarding resource usage~\cite{ayala2020comparison}.

\begin{figure*}[t!]
	\centering
	\includegraphics[scale=0.7]{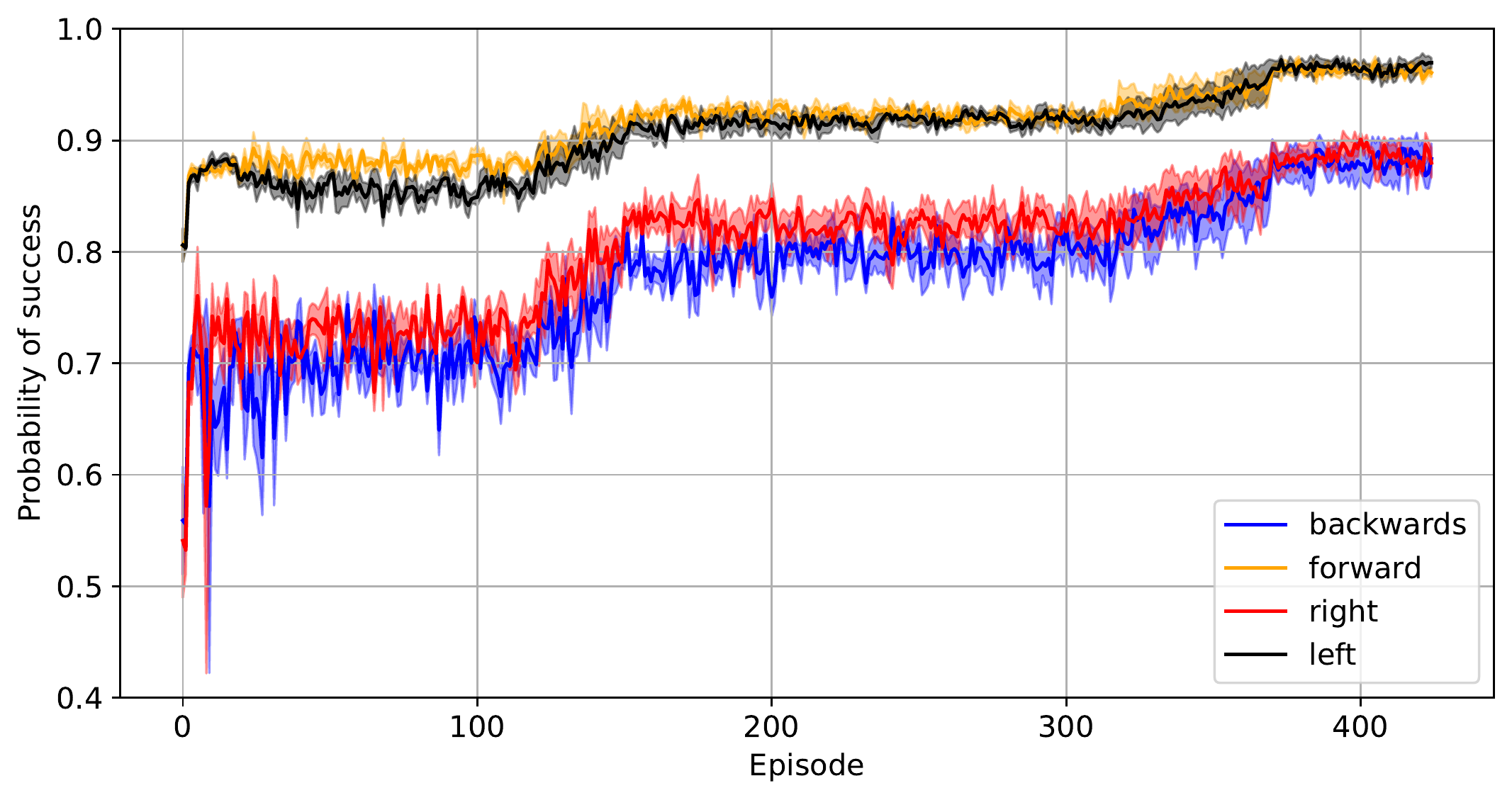}
	\caption[Probabilities for episodic task]{Probabilities of success for every available action of the drone when it is located in a spot close to the bottom-right corner in the episodic scenario.}
	\label{fig:eprob}
\end{figure*}

The robotics scenario built in Webots consists of two objects: the first object is a mailbox. This
mailbox is placed at a random location on a 40~m by 40~m square. The second object is a flying agent, both are
depicted in \cref{fig:sim}. During the experiment, the agent learns how to find
the mailbox as fast as possible. After the flying device reaches a certain height, the agent will take over
the control and can choose between one of four movement options for each time step: move left, move right, go
forward, go backward, as visualized in \cref{fig:movement}. Each movement
option covers a distance of \si{1}{\meter}. The $x$ and $y$ coordinates of the drone as well as
information on the current distance to the mailbox, without specifying the direction towards them,
represent the state. The goal of the agent is to close the distance between the drone and the mailbox,
until a fixed lower boundary has been surpassed and the mailbox is considered to be found. In order to
simplify the training of the agent, the state space is discretized. Movement outside of this predefined
square is restricted as well and will be punished, if the agent attempts to move outside of it.

Based on this setup, an episodic and a non-episodic task are devised. Both of these tasks are then
enhanced with the calculation of the success probabilities.

\subsection{Episodic task}
\label{sub:sc1}
The goal is to find the mailbox as fast as possible. The task is over once the mailbox is found, or once
the limit of 150 time steps has been surpassed. To this end, the reward function is defined as follows:
\\\\\\\\\\
\begin{equation}
r(s)=\begin{cases}
	-0.1 & \parbox[t]{5cm}{per time step}\\
	-100 & \parbox[t]{5cm}{if the time limit is surpassed}\\
	-100 & \parbox[t]{5cm}{if the agent attempts to move outside of the 40m square}\\
	+1  &  \parbox[t]{5cm}{if the distance to the mailbox has decreased}\\
	-1 &  \parbox[t]{5cm}{if the distance to the mailbox has increased}\\
	+100 & \parbox[t]{5cm}{if the mailbox is found}
	\end{cases}
\end{equation}

\subsection{Non-episodic task}
\label{sub:sc2}
For the non-episodic task, the goal is to find as many boxes as possible. Once a mailbox has been
found, the reward will be assigned, the found mailbox is removed and a new mailbox is spawned at a
random location. Since this is a continuous task, there is no terminal state. The rewards are structured
as follows:

\begin{equation}
r(s)=\begin{cases}
	-100 & \parbox[t]{5cm}{if the agent attempts to move outside of the 40m square}\\
	+1  & \parbox[t]{5cm}{if the distance to the mailbox has decreased}\\
	+100 & \parbox[t]{5cm}{if the mailbox is found}
	\end{cases}
\end{equation}

\section{Evaluation}
\label{sec:eval}
To implement the reinforcement learning agent, the DQN algorithm is used, specifically the
implementation provided in Stable~Baselines3~\cite{stable-baselines3}. The environment for the agent is
built with OpenAI~Gym~\cite{gym}. For both tasks, the agent learned over the course of 35000 time steps,
with a learning rate of $0.001$. For each task, five agents were trained consecutively. The displayed
results are averages and standard deviations of the agents within their task group, starting after the
replay buffer was filled and the learning process has begun (after 9750 time steps).

\begin{figure*}[t!]
	\centering
	\includegraphics[scale=0.7]{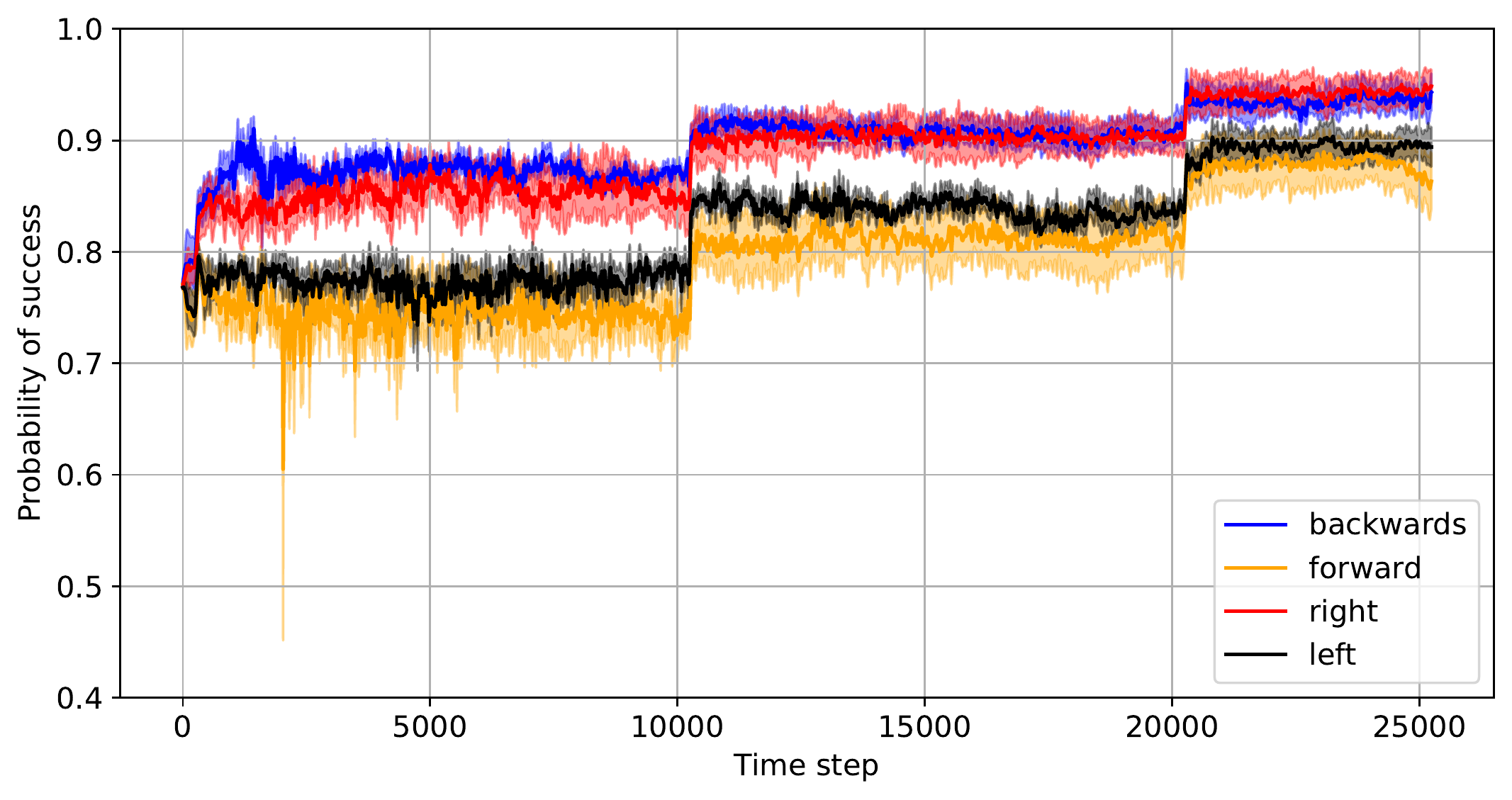}
	\caption[Probabilities for non-episodic task top-left]{Probabilities of success for every available action of the drone when it is located in a spot close to the
	top-left corner in the non-episodic scenario.}
	\label{fig:neprobleft}
\end{figure*}
\begin{figure*}[t!]
	\centering
	\includegraphics[scale=0.7]{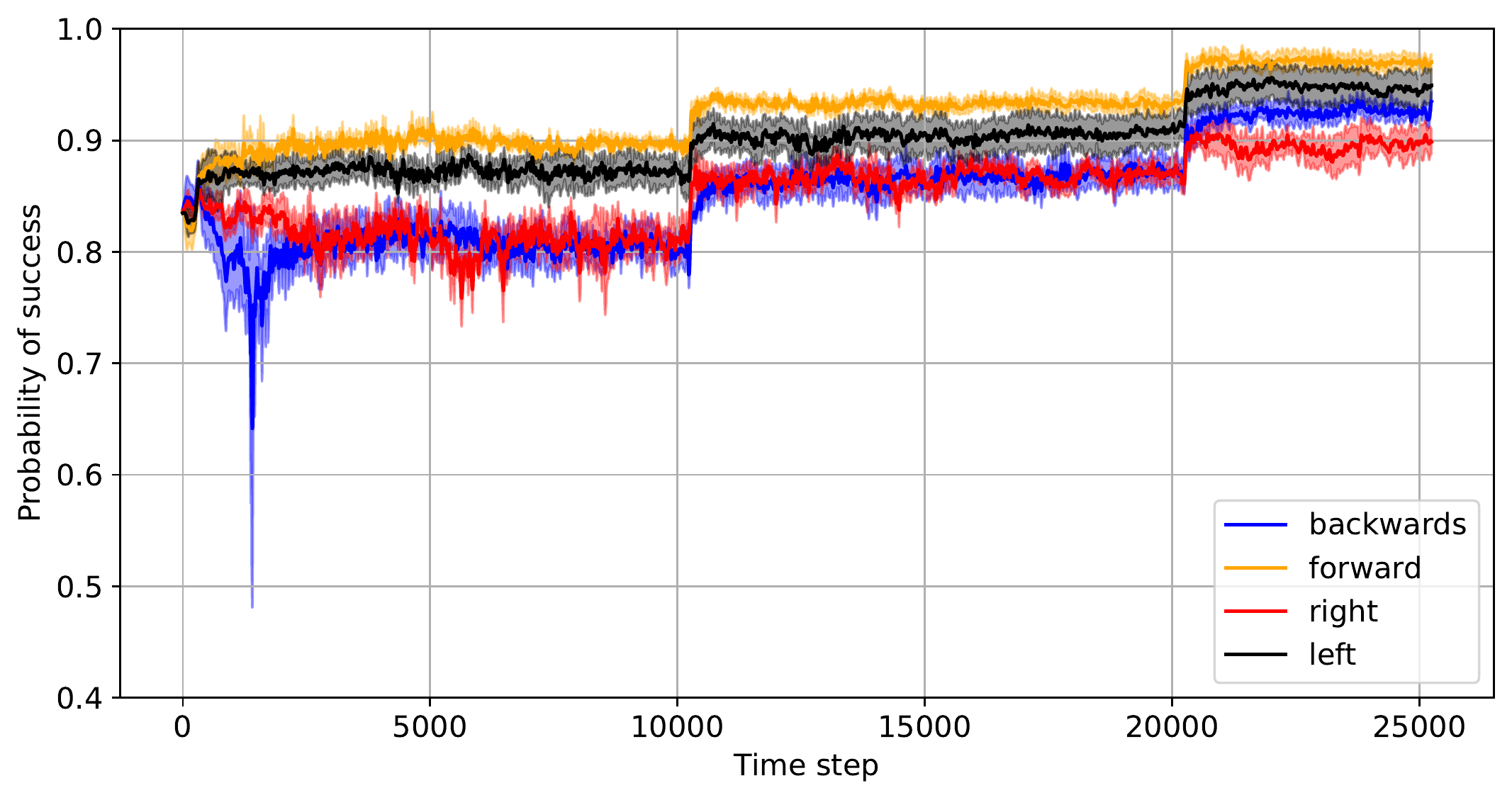}
	\caption[Probabilities for non-episodic task bottom-right]{Probabilities of success for every available action of the drone when it is located in a spot close
	to the bottom-right corner in the non-episodic scenario.}
	\label{fig:neprobright}
\end{figure*}

\subsection{Episodic task}
\label{sub:eval1}
Over the course of training, the Q-values for a specific position were logged after each completed
episode. This position corresponds to a spot close to the bottom-right corner, if the 40m square is
observed from above. Therefore, the actions to move backwards or right will rarely achieve an
improvement with regards to the distance to the mailbox and will instead lead the drone closer to the
aversive region. Therefore, it is assumed that these two actions will have smaller probabilities
attached to them than moving forward or left.

Since some of the Q-values are negative, a transformation of these Q-values is necessary before
\cref{eq2} can be applied. Therefore, the Q-values are normalized to the interval $(0, R^T]$, using
\cref{eq3}. In this equation, $Q_{max}$ and $Q_{min}$ refer to the maximum and minimum respectively of
all observed Q-values for every action $a$ in a particular state $s$, whereas $b$ refers to a small
non-zero constant, leading to a normalization to the previously specified interval on a per-state basis.
After that, the probabilities can be calculated for each episode. The results are displayed in
\cref{fig:eprob}.

\begin{equation}\label{eq3}
Q_{norm}(s,a)=\frac{(Q(s,a)-Q_{min})(R^M - b)}{Q_{max}-Q_{min}}
\end{equation}

The probabilities for moving backwards or right are close together over the course of training and
converge to a similar value as well, just as the success probabilities for moving left or forward.
However, moving backwards or right indeed has a lower success probability attached to it, as
hypothesized above, though the absolute values are still high. The magnitude of these probabilities can
likely be attributed to the simplicity of the task as well as the fact that the episode does not end if
the next movement action is either backwards or right.

These probabilities could now be used to generate explanations with a certain template. For example, if
someone asked why the agent moved left from the bottom-right corner instead of right, it is able to
answer, following a template like “I moved left because it has a success probability of 97~\%, whereas
moving right only has a success probability of 88~\%”.

\subsection{Non-episodic task}
\label{sub:eval2}
For the non-episodic task, the Q-values for the same spot near the bottom-right corner are investigated.
In addition to these, the values for a spot near the top-left corner are examined as well. It is
expected that the success probabilities for the four actions behave inversely for each corner, as the
mailbox will, on average, spawn further away from the respective corner.

As with the previous episodic case, some of the resulting Q-values are negative at some point during
training, therefore a similar normalization is applied before utilizing \cref{eq2} to calculate the
success probabilities displayed in \cref{fig:neprobleft} and \cref{fig:neprobright}. In accordance with
the difference between the episodic and non-episodic maximum reward in the denominator, the Q-values are
normalized to the interval $(0, R^S]$ instead of $(0, R^T]$.

For the top-left corner probabilities in \cref{fig:neprobleft}, a result similar to the episodic task is
observed, albeit inverted. The actions for moving backwards and to the right have higher probabilities
of success attached to them over the course of learning, compared to the movement actions for going
left or forward.

The probabilities of success for the movement actions from the bottom-right corner, displayed in
\cref{fig:neprobright}, also behave similar to the observed probabilities in \cref{fig:eprob}. Going
forward has the highest probability of success at around 97~\%, followed by going left at around 95~\%.
The other two movement actions are situated below once again. Therefore, the hypothesis that was set up
at the start of this subsection appears to be true once again. As with the episodic task, the overall
high probabilities can be attributed to the simplicity of the task.

Once again, these probabilities could be used to generate explanations with the previously seen
template. If someone asked why the agent moved right from the top-left corner instead of forward, the
agent could answer “I moved right because it has a success probability of 95~\%, whereas moving forward
only has a success probability of 86~\%” or it could provide other human-like reasoning that follows a different pattern, while still utilizing these probabilities.

\section{Conclusion}
\label{sec:concl}
In this work, an introspection-based approach to generating explanations for the action of RL agents was further evaluated, based on both an episodic and a non-episodic task in which a flying agent learned to find an object at different random positions. 
Results indicate that the introspection-based approach is useful to generate probabilities to express the chance of success per action, both in the episodic and non-episodic robotics task.
Therefore, generating explanations from these values of the form “I have chosen action X over action Y, because action X has a success probability of P” is possible and viable for said robotics tasks. 
In order to enhance the applicability of the introspection-based approach, normalizations with regard to the terminal and maximum reward respectively were introduced, enabling the approach to be used with negative Q-values and large rewards in comparison to the Q-values, further loosening requirements on the Q-values for the application of the introspection approach.\\
\indent The normalization is not without certain flaws though. 
If the agent was not able to learn the task properly, the highest Q-values might still result in large success probabilities, which will not reflect the inability of the agent to solve the specific task. 
Therefore, future work should focus on alleviating this aspect of normalization. 
It has also not been measured to what extent such explanations enhance the trust of users in robotic systems. 
In order to evaluate this, a user study could be conducted in a real-world scenario to quantify the effects.

%% This section was initially prepared using BibTeX.  The .bbl file was
%% placed here later
\bibliography{publications}
\balance
\bibliographystyle{named}

\end{document}